\documentclass[conference]{IEEEtran}
\IEEEoverridecommandlockouts
\usepackage{cite}
\usepackage{amsmath,amssymb,amsfonts}
\usepackage{algorithmic}
\usepackage{graphicx}
\usepackage{textcomp}
\usepackage{xcolor}
\usepackage{marvosym}
\def\BibTeX{{\rm B\kern-.05em{\sc i\kern-.025em b}\kern-.08em
    T\kern-.1667em\lower.7ex\hbox{E}\kern-.125emX}}
\DeclareRobustCommand*{\IEEEauthorrefmark}[1]{%
    \raisebox{0pt}[0pt][0pt]{\textsuperscript{\footnotesize\ensuremath{#1}}}}
\begin{document}

\title{Audio-driven High-resolution Seamless Talking Head Video Editing via StyleGAN}

\author{\IEEEauthorblockN{Jiacheng Su\IEEEauthorrefmark{1}\IEEEauthorrefmark{,}\IEEEauthorrefmark{2},
Kunhong Liu\IEEEauthorrefmark{1}\IEEEauthorrefmark{,}\IEEEauthorrefmark{2}\textsuperscript{\Letter},
Liyan Chen\IEEEauthorrefmark{1}\IEEEauthorrefmark{,}\IEEEauthorrefmark{2}\textsuperscript{\Letter},
Junfeng Yao\IEEEauthorrefmark{1}\IEEEauthorrefmark{,}\IEEEauthorrefmark{2}, 
Qingsong Liu\IEEEauthorrefmark{3},
Dongdong Lv\IEEEauthorrefmark{3}}
\IEEEauthorblockA{\IEEEauthorrefmark{1}School of Film, Xiamen University, Xiamen, China}
\IEEEauthorblockA{\IEEEauthorrefmark{2}Key laboratory of Digital Protection and Intelligent Processing of lntangible Cultural Heritage \\
of Fujian and Taiwan, Ministry of Culture and Tourism, China}
\IEEEauthorblockA{\IEEEauthorrefmark{3}Xiamen Unisound Intelligence Technology Co.,Ltd., Xiamen, China}
\IEEEauthorblockA{sujiacheng@stu.xmu.edu.cn, \{lkhqz, chenliyan, yao0010\}@xmu.edu.cn, \{liuqingsong, lvdongdong\}@unisound.com}
\thanks{\Letter \ Corresponding authors.
}}

\maketitle

\begin{abstract}
The existing methods for audio-driven talking head video editing have the limitations of poor visual effects. This paper tries to tackle this problem through editing talking face images seamless with different emotions based on two modules: (1) an audio-to-landmark module, consisting of the CrossReconstructed Emotion Disentanglement and an alignment network module. It bridges the gap between speech and facial motions by predicting corresponding emotional landmarks from speech; (2) a landmark-based editing module edits face videos via StyleGAN. It aims to generate the seamless edited video consisting of the emotion and content components from the input audio. Extensive experiments confirm that compared with state-of-the-arts methods, our method provides high-resolution videos with high visual quality. 
\end{abstract}


\section{Introduction}
\label{sec:intro}
The audio-driven talking head video editing is an important research topic in the AIGC field, with the aim to generate high-quality talking head video featuring the target individual in synchrony with the input audio. This task is widely employed in film dubbing\cite{1976Film}\cite{2022Towards} and digital human technology\cite{JEREMY2013Avatars} to adjust actors’ lip movements and generate lifelike facial animation of the digital human.

Up to now, many researchers had been devoted to the effective methods. However, most of the studies\cite{zhou2020makelttalk}\cite{lu2021live} failed to work well on high-resolution videos, exhibiting noticeable editing traces and blurred effect. \cite{yin2022styleheat} addressed the issue and achieved high-solution talking face generation by resorting to a pre-trained StyleGAN\cite{karras2019style}. However, it tended to result in reconstruction errors due to the information lost in the feature map.

\begin{figure}[t]
        \centering
        \includegraphics[width=0.47\textwidth]{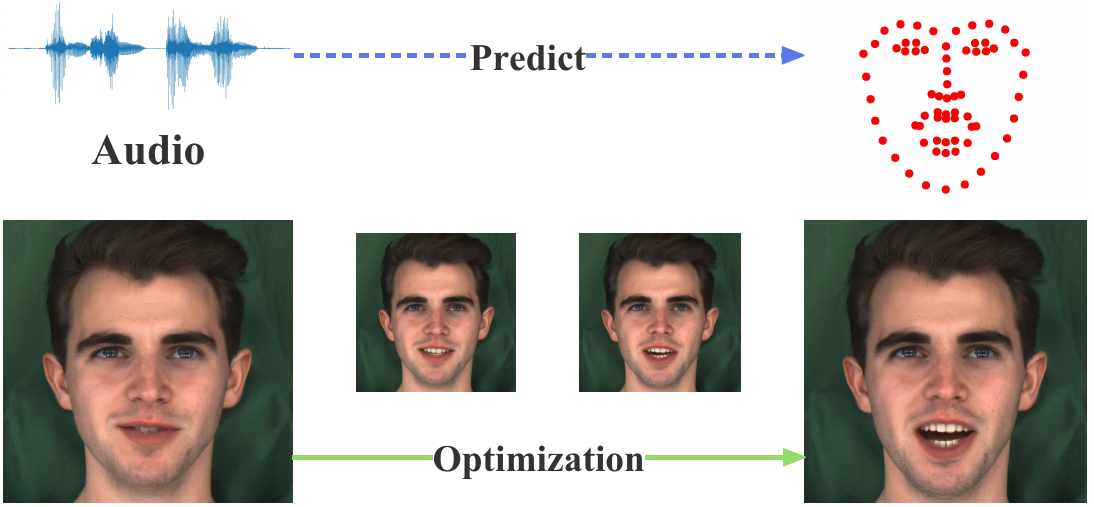}
        \caption{Our method fits the generated frame to the landmark predicted from the given audio.}
        \label{Fig.1}
\end{figure}

Inspired by the previous StyleGAN-based talking head video generation approach\cite{yin2022styleheat}, we propose a novel framework for synchronized facial video editing using StyleGAN to achieve a better visual quality. Instead of feature map, we conduct the editing in the dimension of $\mathcal{W}+$ latent space, which is highly-isentangled for the facial attribute editing. We predict landmarks through an Audio-to-Landmark (AL) module including the Cross-Reconstructed Emotion Disentanglement\cite{ji2021audio} to convey the emotions embedded in the audio and an alignment module to align the head pose. And then an optimization algorithm is proposed to edit frames under the supervision of facial landmarks as shown in Fig. \ref{Fig.1}. Furthermore, the StyleGAN-based editing ensures the high-resolution video, along with a tuning approach\cite{tzaban2022stitch} to make the video seamless. Extensive experiments on two datasets demonstrate the superior performance of our method in talking head video editing.

Our main contributions are summarized as follows:
\begin{itemize}
  \item We propose a novel framework based on StyleGAN for talking head video editing. It enables high-resolution synchronized generation and seamless editing, generating different expressions in accordance with the emotion embedded in the input audio.
  \item We introduce an optimization algorithm to achieve the generation of facial-edited videos via StyleGAN under the supervision of facial landmarks. It simultaneously maintains the identity of the original video characters and the smoothness of the video.
  \item We develop an Audio-to-Landmark module that can generate emotional, pose-aligned facial landmarks corresponding to the target person speaking in the audio. An effective alignment module with the Cross-Attention mechanism is designed to faciliate this process.
\end{itemize}

\begin{figure*}[t]
        \centering
        \includegraphics[width=1.0\textwidth]{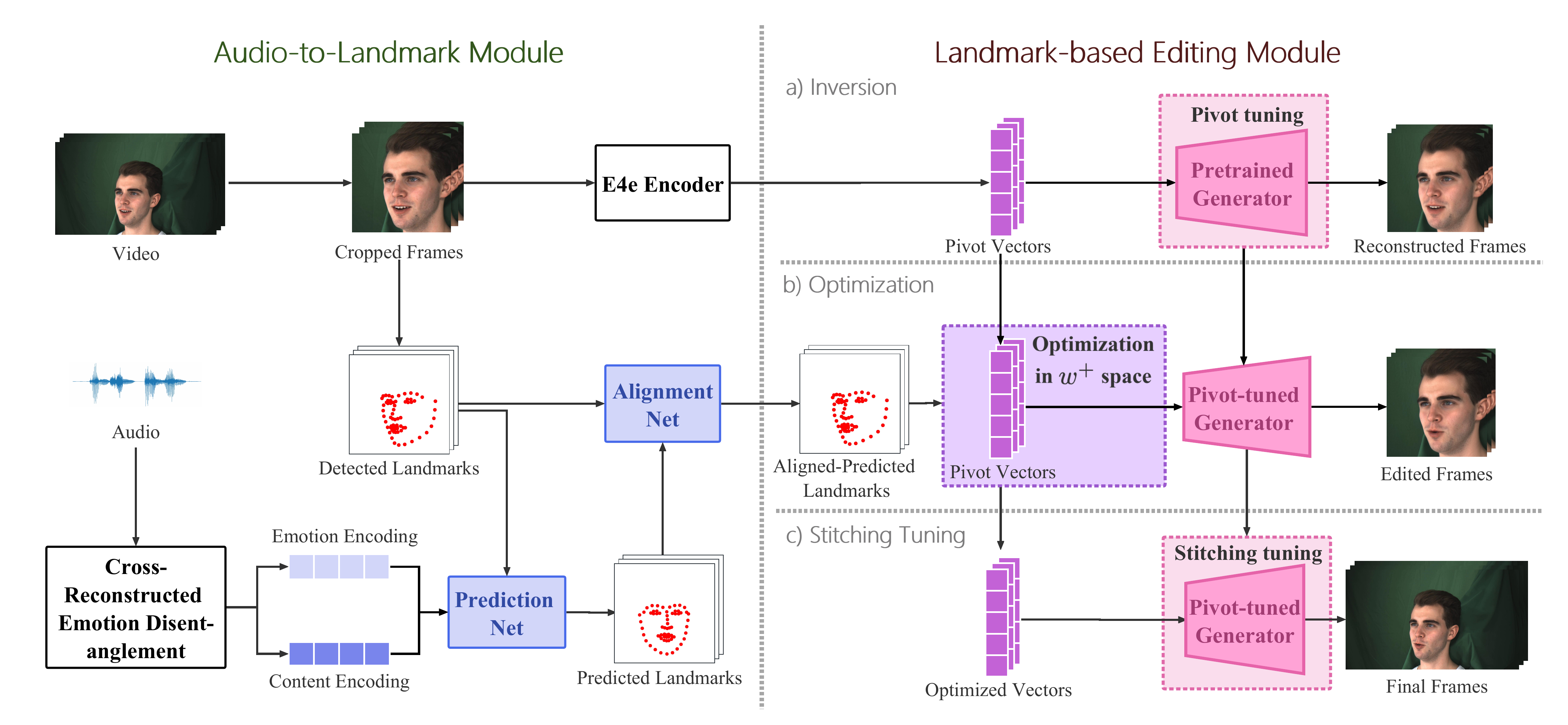}
        \caption{\textbf{The framework of our method.} Our method is divided into two parts: (1) Audio-to-Landmark Module; (2) Landmark-based Editing Module, which contains three steps: a) Inversion, b) Optimization, c) Stitching Tuning.}
        \label{Fig.2}
\end{figure*}

\section{related work}
\textbf{Talking head video editing.} Talking Head Video Editing encompasses a spectrum of approaches\cite{ji2021audio} leveraging generative models like GANs\cite{heusel2017gans}\cite{harkonen2020ganspace} and VAEs\cite{kingma2013auto} for video manipulation. Research\cite{zhang2021facial} explores conditional GANs for specific attribute editing, including facial reenactment, image-to-image translation, and 3D reconstruction\cite{yao2022dfa}\cite{guo2021ad}. Techniques involving facial landmark manipulation, style and motion transfer, temporal consistency, and real-time interactive tools are prominent\cite{yin2022styleheat}. Advancements focus on neural rendering and other novel methods ensuring realistic synthesis\cite{mildenhall2021nerf}. These endeavors collectively drive innovations in video editing, offering capabilities for nuanced facial expressions, pose adjustments, and compelling visual transformations in videos. Compared to them, our method goes a step further by increasing the resolution of the video and enabling seamless editing of the video.

\textbf{StyleGAN.} StyleGAN\cite{karras2019style} had garnered significant interest due to its capacity of producing high-fidelity facial images and its notably disentangled feature space. Consequently, the emergence of StyleGAN editing through GAN inversion\cite{goodfellow2020generative} gained popularity within swiftly evolving GAN landscape. GAN inversion techniques manipulated images by traversing the latent space of pretrained models\cite{gal2022stylegan}\cite{fox2021stylevideogan}. These methods were broadly categorized into optimization-based\cite{abdal2019image2stylegan}\cite{abdal2020image2stylegan++}\cite{abdal2021styleflow}, encoder-based\cite{richardson2021encoding}, and hybrid approaches\cite{chen2019hierarchical}. Among these methods, optimization-based approaches achieved superior reconstruction quality, although they required per-image optimization. In talking head video editing, \cite{yin2022styleheat} applied StyleGAN combined with optical flow to this task. However, due to the limitation of the inversion method, it failed to present good visual effects. In this paper, we design an optimization-based approach in pursuit of a better visual effect.

\section{method}
    The workflow of our mothod is illustrated in Fig. \ref{Fig.2}, consisting of the Audio-to-landmark (AL) module and Landmark-based Editing (LE) module.

    \subsection{The Audio-to-Landmark Module}
    The AL module provides facial landmarks for optimization, including emotion disentanglement, prediction and alignment.
        
        \textbf{Emotion Disentanglement and Prediction.} We adopt the cross-reconstructed emotion disentanglement to extract the emotion and content components from the audio. We follow the approach in \cite{ji2021audio} to establish training pairs, and exchange their emotion embedding and content embedding for the cross-reconstruction. It effectively decouples the audio signal into two separate representations via subjecting the cross-reconstructed results for training. Given the emotion embedding and content embedding, the AL module predicts the landmark displacements by a long short term memory (LSTM) network followed by a two-layer MLP as the prediction network.

        \textbf{Alignment.} It is noted that the pose information of the face is not available in the audio, so we design an alignment network to impose pose constraints on the predicted landmarks with the Cross-Attention ($CA$) mechanism. Here the token of the predicted facial landmarks $x_{\text{pred}}$ serves as the query, and the token of the original facial landmarks $x_{\text{orig}}$ acts as both key and value. This mechanism allows the model to capture the interdependencies between the predicted and original landmarks, expressed as:
        \begin{equation}
        \begin{gathered}
            CA(x_{\text{pred}},x_{\text{orig}}) = \text{softmax}\left(\frac{QK^T}{\sqrt{t}}\right)V, \\
         where Q=x_{\text{pred}}W^Q, \quad K=x_{\text{orig}}W^K, \quad V=x_{\text{orig}}W^V.
        \end{gathered}
        \end{equation}
        Here, $W^Q$, $W^K$, $W^V$ denote the corresponding projection matrices, and $t$ denotes the size of the token. We utilize a dense layer to output the ultimate result.

        \textbf{Training.} We train our networks with the MEAD dataset\cite{wang2020mead}, which comprises a substantial collection of speech videos featuring multiple angles and emotions. Two videos of the same speech segment captured from different angles serve as a training pair. We extract landmarks $l_p$, $l_f$ from two videos. $l_p$ provides pose information and serves as the alignment ground truth, and $l_f$ provides facial information. Both sets of landmarks are fed into the Alignment Network($AN$) to obtain aligned landmarks $l_\text{aligned}$. Furthermore, we perform warping on $l_p$ to prevent the leakage of facial information from the original video. The process of alignment is given by:
        \begin{equation}
        l_\text{aligned}=AN(warping(l_p),l_f).
        \end{equation}
        And we calculate the loss by measuring the distance between the aligned landmarks and the landmarks from the original video, defined by:
        \begin{equation}
        L_\text{align}=\|l_\text{aligned} - l_p\|^2_2.
        \end{equation}
        Then, we freeze $AN$ and train the prediction network. Similarly, to prevent information leakage, we distort the input landmarks and calculate the loss by measuring the distance between the predicted aligned landmarks and the landmarks of the input video.

    \subsection{Landmark-based Editing Module}
        The Landmark-based Editing (LE) module generates the edited video via StyleGAN under the supervision of the previous predicted landmarks. To modify frames following the landmarks in a temporal coherent manner, we design a pipeline containing inversion, optimization, and stitching tuning, to achieve a seamless and realistic editing.

        \textbf{Inversion.} To edit the video effectively, we invert each frame into the latent space of the GAN, as the effectiveness of the inversion reconstruction directly affects the realism and quality of the generated video. PTI\cite{roich2022pivotal} is used to tune the generator, to reconstruct the original frame in a more editable region.

        Given a sequence of cropped and aligned video frames $\sum_{i=1}^Nf_i$ where $N$ denotes the number of the frames, we invert each frame $f_i$ using an E4e Encoder\cite{tov2021designing} to obtain the corresponding latent vector $w_i$. These latent vectors are used as 'pivots' for PTI. The reconstructed image $r_i$ is generated by a generator $G$ with the parameter $\theta$, where $r_i=G(w_i;\theta)$. The objective of PTI is set as: 
        
        \begin{equation}
        \mathop{\min}_{\theta}\frac{1}{N}\sum_{i=1}^N(L_\text{LPIPS}(f_i,r_i)+\lambda_{L_2}^{P}{L_2}(f_i,r_i)+\lambda_{R}^{P}L_R),
        \end{equation}
        where $L_\text{LPIPS}$ denotes the LPIPS perceptual loss proposed by \cite{zhang2018unreasonable}, ${L_2}$ denotes the pixel-wise MSE distance, and ${L_R}$ denotes the locality regularization described by \cite{roich2022pivotal}. $\lambda_{L_2}^{P}$ and $\lambda_{R}^{P}$  are constants across all experiments.

        \textbf{Optimization.} Our optimization goal is to edit the target face in video in synchrony with the predicted landmarks. We design a multiple loss function to keep a balance between synchronized editing, identity preservation and video quality. For each frame $f_i$, we optimize the latent vector $w_i$ by minimizing the loss made up of three terms between the original frame $f_i$ and the generated frame $x_i$, where $x_i=f(w_i )$ and $f(\cdot)$ is the pivot-tuned generator. The loss function $L_\text{loss}$ is calculated as:

        \begin{equation}
        L_\text{loss}=\lambda_{\text{LPIPS}}L_\text{LPIPS}+\lambda_{\text{fan}}L_\text{fan}+\lambda_{\text{smooth}}L_\text{smooth}.
        \end{equation}
        Here, $L_\text{LPIPS}$, $L_\text{fan}$, $L_\text{smooth}$ are used to optimize the input variable $w_i$ through gradient descent, with keeping the network weights fixed. $\lambda_{\text{LPIPS}}$, $\lambda_{\text{fan}}$, $\lambda_{\text{smooth}}$ are constants across all experiments.
        
        $L_\text{LPIPS}$ is calculated on Learned Perceptual Image Patch Similarity (LPIPS) by comparing two input images in a deep feature space, so as to measure the perceptual similarity between the generated images and the original frames. 

\begin{figure*}[t]
        \centering
        \includegraphics[width=1\textwidth]{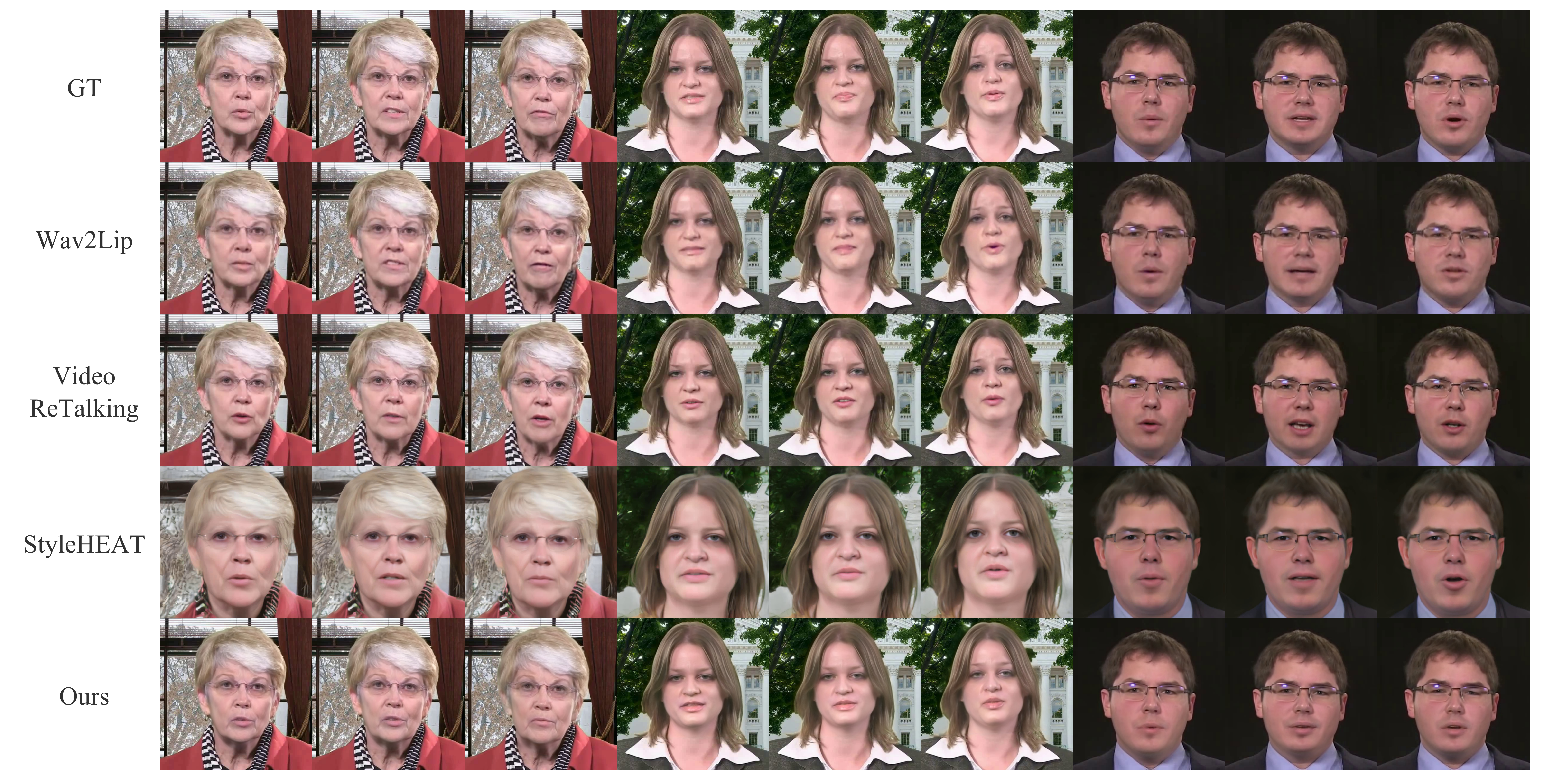}
        \caption{\textbf{Qualitative comparisons with the state-of-the-art methods.} Three examples with different speech content in HDTF dataset, comparing with Wav2Lip, VideoReTalking, and StyleHEAT.}
        \label{Fig.3}
\end{figure*}

        $L_\text{fan}$ measures the divergence between the facial landmarks of the generated images and the target facial landmarks. However, the facial landmarks are gradient-free in practice. So a landmark heat map extraction model, FAN\cite{bulat2017far}, is deployed to solve the problem with generating a three-dimensional matrix $H\in{R^{64\times64\times{n}}}$, consisting of $n$ heat maps of $64\times64$. Here $n$ denotes the number of landmark points. Specially, given two images $I^1$ and $I^2$, the corresponding heat maps matrix are represented as $H^1=FAN(I^1)$ and $H^2=FAN(I^2)$. For each landmark point $P_i$, let $d(H^1_{P_i},H^2_{P_i})=\sqrt{(H^1_{P_i}-H^2_{P_i})^2}$ denote the distance between the heat maps. It should be noted that the landmark points in the mouth and eyes regions are key to the fluency of generated videos, so we set a biased weight for the heat map of each landmark point when calculating the distance. The biased weight vector is defined by $\lambda_{\text{landmark}}=\sum_{i=1}^n{a_i}$, which contains weight $a_i$ for the i-th heatmap. This vector is adjusted to generate more consistent facial features, and the landmark loss between two images is calculated by:

        \begin{equation}
        L_\text{fan}(f_i,x_i,\lambda_\text{landmark})=\sum_{i=1}^n{a_i}\sqrt{(H^1_{P_i}-H^2_{P_i})^2}.
        \end{equation}

        $\lambda_{\text{smooth}}$ measures the similarity between the generated consecutive frames. Typically, in the video generation, the continuity is evaluated by computing distance losses between consecutive  frames. Unlike traditional loss functions in StyleGAN-based generation method, we restrict the magnitude of changes between consecutive frames by the distance in latent space, so as to ensure video continuity by preserving coherence in appearance across frames. The smoothness loss is given by:

        \begin{equation}
        L_\text{smooth}(w_{i-1},w_i)=\|w_{i-1}-w_i\|^2_2.
        \end{equation}

        \textbf{Stitching tuning.} To seamlessly blend the generated frames with the original frames, we adjust the generator with stitching tuning[23]. Given the original frame $f_i$, we first use off-the-shelf pretrained segmentation network\cite{yu2021bisenet} to produce segmentation masks $m_i$. Then, we expand the edge of the segmentation mask, considering the expanded region as the boundary region $b_i$. It employs L1 loss to compute the region loss $L_r$ as the weighted sum of two losses: the boundary loss ($L_{L1}$) and the mask region($\lambda_{m}L_{L1}$). We optimize the weights $\theta$ of the generator jointly by:

        \begin{equation}
        \begin{gathered}
        L_r=L_{L1}(G(w_i;\theta)\odot{b_i},f_i\odot{b_i})+ \\
        \lambda_{m}L_{L1}(G(w_i;\theta)\odot{m_i},G(w_i;\theta_\text{orig})\odot{m_i})
        \end{gathered}
        \end{equation}
        Here, $G$ denotes the generator, $\theta_\text{orig}$ denotes the original weights of $G$, $\odot$ is the element-wise multiplication and $\lambda_m$ is constant across all experiments. The boundary loss guides better alignment of the edited segment's boundary with the original frame, and the mask region loss is designed to provide masks to the edited segments excluding boundary.

\begin{table*}[h]
\centering
\scalebox{1.215}{
\renewcommand{\arraystretch}{2}
\begin{tabular}{|l|c|c|c|c|c|c|c|c|c|c|}
\hline
Method/Score & \multicolumn{5}{c|}{MEAD} & \multicolumn{5}{c|}{HDTF} \\ 
\cline{2-11}
& FID$\downarrow$ & PSNR$\uparrow$ & SSIM$\uparrow$ & LPIPS$\downarrow$ & F-LD$\downarrow$ & FID$\downarrow$ & PSNR$\uparrow$ & SSIM$\uparrow$ & LPIPS$\downarrow$ & F-LD$\downarrow$ \\ 
\hline
Wav2Lip & 17.23 & 34.08 & \textbf{0.9702} & 0.1110 & 2.973 & 14.92 & 34.82 & \textbf{0.9664} & 0.0839 & \textbf{2.861} \\ 
VideoReTalking & 17.89 & 34.10 & 0.9656 & 0.1237 & 4.045 & 14.09 & 31.07 & 0.9411 & \textbf{0.0783} & 3.879 \\ 
StyleHEAT & - & - & - & - & - & 18.02 & 31.21 & 0.6019 & 0.1729 & 4.353 \\ 
Ours & \textbf{13.74} & \textbf{34.91} & 0.9677 & \textbf{0.0715} & \textbf{2.633} & \textbf{13.85} & \textbf{35.69} & 0.9212 & 0.0792 & 3.411 \\ 
\hline
\end{tabular}}
\\ [0.2cm]
\caption{\textbf{Quantitative comparisons with the state-of-the-art methods.} We calculate the landmark accuracy and video qualities of different methods.}
\label{tab:combined}
\end{table*}

\section{experiments}

    \subsection{Datasets and Implementation Details}Two standard datasets, MEAD\cite{wang2020mead} and HDTF\cite{zhang2021flow}, are adopted to evaluate the performance of different methods. Both datasets are high-resolution audio-visual dataset. In addition, MEAD offers multi-emotional talking head videos for the demonstration of generating different emotions. In experiments, $\lambda_\text{LPIPS}=1$, $\lambda_\text{fan}=5e^{-3}$ and $\lambda_\text{smooth}=1e^{-4}$. We use a learning rate of $1e^{-3}$ and optimize each frame 300 iterations. We follow PTI\cite{roich2022pivotal} and STIT\cite{tzaban2022stitch} respectively for parameter settings.

\begin{figure}[ht]
        \centering
        \includegraphics[width=0.48\textwidth]{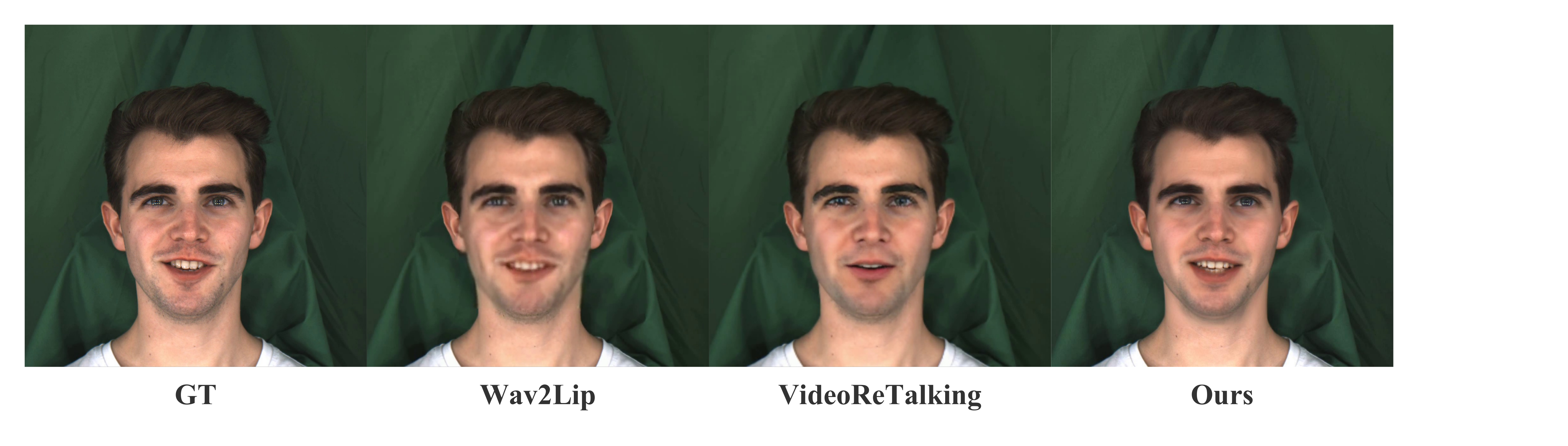}
        \caption{More results in MEAD dataset comparing with Wav2Lip\cite{prajwal2020lip}, VideoReTalking\cite{cheng2022videoretalking}.}
        \label{Fig.4}
\end{figure}

    \subsection{Experimental Result}We compare our work with three models: Wav2Lip\cite{prajwal2020lip}, VideoReTalking\cite{cheng2022videoretalking}, StyleHEAT\cite{yin2022styleheat}. As the state-of-art methods in the task of talking head video, Wav2Lip shows well lip sync capability, and VideoReTalking produces high-resolution edited video. StyleHEAT is selected as it is also designed based on StyleGAN, and the first frame of the target video is used as its inputsin expriments.  More results are provided in supplementary materials.

    \textbf{Qualitative Comparisons.} To better illustrate the comparisons with other methods, some generated frames (514$\times$514) by different methods are shown in Fig. \ref{Fig.3} based on the HDTF dataset. Our method generates high-fidelity emotional talking face video, and performs well in lip syncing. Compared to StyleHEAT, our method maintains better identity consistency. In contrast with Wav2Lip and VideoReTalking, our method produces videos with higher visual quality. It's noticeable that Wav2Lip's output appears comparatively blurry, especially with evident alterations around the mouth. While VideoReTalking aims to enhance the quality of generated images, there are imperfections around the eyes. As shown in Fig. \ref{Fig.4} where all the iamges are 1080$\times$1080, these issues become more evident on the higher-resolution MEAD dataset. Instead, our method enables high-resolution generation and seamless editing.

    \textbf{Quantitative Comparisons.} To evaluate the quality of the generated videos, FID\cite{heusel2017gans}, PSNR, SSIM\cite{wang2004image}, LPIPS\cite{zhang2018unreasonable} scores are used in the experiments. We extract facial landmarks from the generated sequences and the ground truth sequences, and the evaluation of facial motions are conducted by Facial Landmark Distance(F-LD). The results are illustrated in Table \ref{tab:combined}. Our method performs well in both video and sync quality. Specially, we observe from the FID and PSNR metrics that our method exhibits consistently high-quality performance in terms of image quality. Due to Wav2Lip's modification being confined to the lower facial region, our method slightly trails behind in the similarity metric SSIM. Note that as is trained on the HDTF dataset, StyleHEAT exhibits significant distortion when tested on MEAD. Therefore, results from StyleHEAT on MEAD is not adopted.
    
\begin{table}[t]
\centering
\resizebox{0.45\textwidth}{!}{%
\renewcommand{\arraystretch}{2}
\begin{tabular}{|l|c|c|c|c|c|}
\hline
Method/Score & FID$\downarrow$ & PSNR$\uparrow$ & SSIM$\uparrow$ & LPIPS$\downarrow$ & F-LD$\downarrow$ \\
\hline
w/o $L_\text{smooth}$ & 17.13 & 29.86 & 0.9341 & 0.1475 & 5.518 \\
with $L_\text{smooth}^F$ & 15.20 & 34.41 & 0.9662 & 0.0724 & 2.890 \\
Ours & \textbf{13.74} & \textbf{34.91} & \textbf{0.9677} & \textbf{0.0715} & \textbf{2.633} \\
\hline
\end{tabular}%
}
\\ [0.2cm]
\caption{\textbf{Quantitative ablation study for smoothness loss.}}
\label{tab 2}
\end{table}

\begin{figure}[ht]
        \centering
        \includegraphics[width=0.48\textwidth]{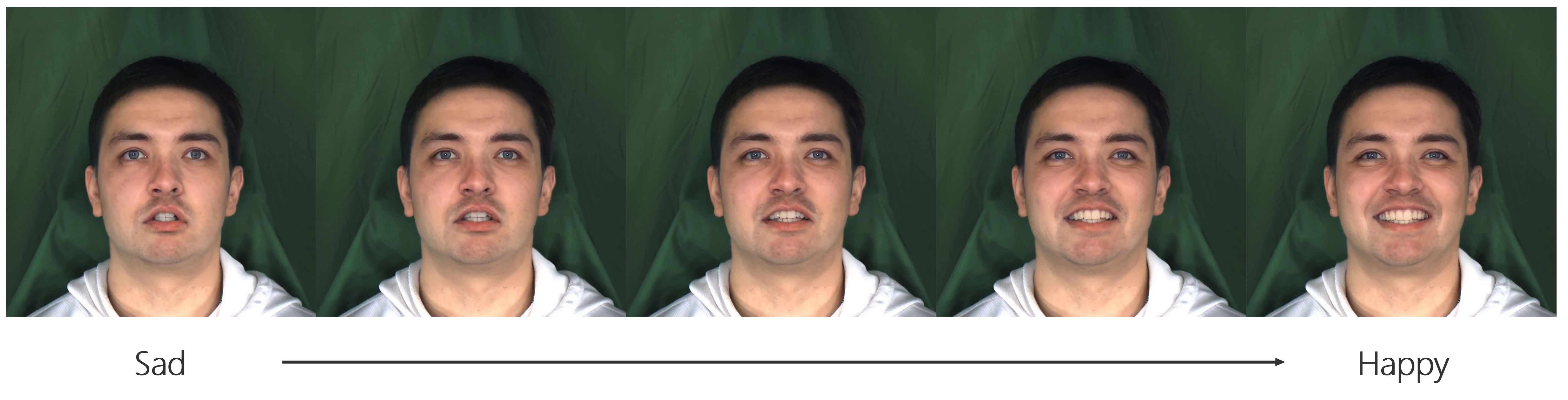}
        \caption{\textbf{Emotional editing.} We show the frames generated with linear emotional Variation.}
        \label{Fig.5}
\end{figure}

    \subsection{Ablation Study}The ablation study is mainly conducted by changing the loss function. Our method calculates the distance between subsequent latent vectors as smoothness loss to keep features more stable, which is not directly constrained by the perceptual loss or the landmark loss over time. From Table \ref{tab 2}, it's obvious that the smoothness loss effectively enhances the quality of the generated videos. We compare the smoothness loss with the loss $L_\text{smooth}^F$, calculating the distance between consecutive frames. And the results confirm that enforcing continuity constraints on vectors in latent space is equally effective for the generated images in the target domain, and even exhibits superior quality.

    \subsection{Emotional editing.}Our method can extract emotional information from the audio to achieve emotional editing. As shown in Fig. \ref{Fig.5}, we generate frames with linear emotional variation. Compared to the widely deployed methods\cite{prajwal2020lip}\cite{cheng2022videoretalking} with modifying only half of the face, our method can better represent character expressions by predicting landmarks for the whole face. The ability to generate high-definition images also allows for a better representation of the details of facial expressions.

\section{CONCLUSION} In this paper, we propose a novel framework for talking head video editing to promote the visual effects with high resolution. The proposed optimization algorithm better fits the facial video to the target facial landmarks. And then an audio-to-landmark module is designed to predict emotional landmarks from audio, with an effective alignment network along with the Cross-Attention mechanism to get the aligned predicted landmarks. Based on these modules, our method produces seamless editing with high-resolution outputs, allowing the generation of different expressions in accordance with the emotion embedded in the input audio. Experiments show that compared with other state-of-the-art methods, our method offers high performance in the generation of high quality video.

\section{ACKNOWLEDGEMENT} This work is supported by National Key R\&D Program of China(2023YFC2604400), Fujian Science and Technology Plan Industry-University-Research Cooperation Project (No.2021H6015), and the public technology service platform project of Xiamen City(No.3502Z20231043).

\bibliographystyle{IEEEtran}
\bibliography{reference}

\end{document}